\documentstyle[a4]{article}
\title{\bf
How much has information technology contributed to linguistics?}
\author{
Karen Sparck Jones\\
Computer Laboratory, University of Cambridge\\
New Museums Site, Pembroke Street, Cambridge CB2 3QG, England\\
{\em ksj@cl.cam.ac.uk}}
\date{December 1996}

\begin{document}
\maketitle

\begin{center}
To appear in\\
{\em Information technology and scholarly disciplines}, Ed J.T. Coppock\\
Proceedings of a British Academy Symposium, London: The British Academy
\end{center}

\vspace{10mm}

\begin{abstract}

Information technology should have much to offer linguistics, not
only through the opportunities offered by large-scale data analysis
and the stimulus to develop formal computational models, but through
the chance to {\em use} language in systems for automatic natural language
processing. The paper discusses these possibilities in
detail, and then examines the actual work that has been done.
It is evident that this has so far been primarily research within
a new field, computational linguistics, which is largely motivated by
the demands, and interest, of practical processing systems, and that
information technology has had rather little influence on linguistics
at large. There are different reasons for this, and not all good ones:
information technology deserves more attention from linguists.

\end{abstract}

\section{Introduction}

   There are two potential roles for information technology (IT) in
linguistics, just as in other areas: as a means of developing and testing
models and as a means of gathering and analysing data.
For example, one may use a computer to help make some model of word
formation properly specific, and also to gather and analyse some data on
word forms.
Linguistics thus has the same types of use and benefit for computing
as other academic areas, like archaeology or economics.

   IT in linguistics can give both of these a sharper edge. Thus in the
lesser case, data analysis, we can use the machine not merely to interpret
data but to gather it.  In the archaeological case, we can analyse
supplied descriptions of pots to hypothesize a typological sequence, say, but
the descriptions have to be supplied. Even with such aids as automated
image analysis, the human input required is generally large. In the language
case, in contrast, if we want to determine lexical fields, we can just pull
text off the 
World Wide Web. We still need humans to supply the classification theory,
but one can't get everything from nothing, and the detailed human work is
much less than in the archaeological case.

   But much more importantly in relation to modelling, in the language case
we can not only use computers to develop and test models, in the normal way.
We can apply computers operationally, and hence creatively, for the very same
language-using
tasks as humans do. For example, if we have a model of speech
production, we can build a speech synthesizer which can be attached to
an advice system to generate new utterances in response to new inquiries,
now. Again, with a translation system based on some model of translation, we
can actually exercise this model, in an especially compelling way, by
engaging in translation. But however good our archaeological models of the
spread of neolithic agriculture are, they cannot go out and plough up
untilled land at the rate of so and so many yards a day,

   It is this productive new, i.e. {\em real}, application of computational
models that makes interaction between IT and linguistics interesting, in
the same way that the interaction embodied in biotechnology is.
Model validation, with its supporting need for
serious data, is a good reason for examining what may be called the IT
technology push into linguistics. But the potentially productive use, in
practical applications, {\em for} models, and the especially strong
validation this implies, means that IT technology pull from linguistics
can also be assessed for what it has contributed to linguistics.

   This is all an exciting idea;
and it has stimulated a wholly new research field, Computational
Linguistics. But IT has nevertheless had much less influence on linguistics
in general than one would expect from the fact that words,
the stuff of language, are now the pabulum of the networks and figure
more largely in what computers push around than numbers do:
computational linguistics remains a quite isolated area within
linguistics. Linguistics has
also had far less influence than might be expected on task systems that process
natural language.
\footnote{In computing it is necessary to distinguish natural language
from programming language.}
I believe there are both good and bad
reasons for this state of affairs, and will consider these after looking
in more detail at specific forms of possible, and actual, interaction
between linguistics and IT.

\subsection*{Exclusions}

   The range of specific areas to examine is large. I shall exclude
two that, however intellectually important to their communities, or
practically valuable, I see as peripheral to my main topic.

   One is the whole area labelled `computers and the humanities', when this
deals with language data for specific individuals or sources, considered
in relation to author attribution or manuscript genealogies, say,
or in content
analysis as in the study of
the way political terms are used in newspapers. This is
where all of the utilities exemplified by SGML have a valuable role in
supporting scholarship (see e.g. Sperberg-McQueen (1994) on the Text
Encoding Initiative); as illustrative titles for applications of this kind
we can take such random examples from the ALLC-ACH '96 Bergen Conference
as `The Thesaurus of Old English' database: a research tool for historians
of language and culture'; ` ``So violent a metaphor.'' Adam Smith's
metaphorical language in the {\em Wealth of Nations}'; and `Book, body and
text: the Women Writers Project and problems of text encoding'.   
But I am excluding this type of work as itself on the borderline of
linguistics.

   The other major area I shall exclude is language teaching. Again, IT
already has an established role in this, though far more as a dumb waiter
than as an intelligent tutor that continuously adapts the content and
presentation
of lessons to the individual student. So far, there has been little
progress in the development of teaching programs that would de facto
constitute a serious test of alternative accounts of grammar or choose
among performance models of language processing.

   I am also no expert on speech research, so will only note some
`place-holding'
points on spoken as opposed to written language.

   I shall however, for the moment, take the scope and style of
linguistics as properly large, and not restrict linguistics as an area
of endeavour or discipline to a particular purpose or stance. I
shall return to the consequences of contemporary attitudes to these later.

\subsection*{Structure}

   I will start by considering what IT can in principle (but also
soberly) offer linguistics. I will then assess how far linguists have
exploited IT in practice. Finally, I will try to explain the
present state of affairs. The focus is on the contribution of IT
to linguistics, so I shall not attempt a systematic treatment of
the work done, in natural language processing (NLP), by those who
do not think of themselves as linguists,
as opposed to engineers, or consider, in detail,
the influence of linguistics on this work. I will, however, refer to both
of these as this is necessary to round out my main argument.

   I have identified two main roles for IT in linguistics: data gathering
and modelling. Of course these come together when corpus data is used to
test some theory. However there are in general marked differences between
those who cut the corn and those who sharpen the sickles. I shall
therefore consider first work with data, and then the development of theory.

\section{IT possibilities}

\subsection*{Data work}

   Data, or corpus, work is a natural for IT: computers can so rapidly
and painlessly match, sort, count and so forth vast volumes of material;
and as these are increasingly text that is already machine-readable,
so there is no data-entry effort for the linguist, IT would appear now
to have much to offer. The points below refer primarily to natural,
independently-produced text rather than to elicited data, though they
also apply to the latter; and automatic manipulation of data
can also of course be useful for material marked up by the linguist.

   Corpus work is of value at three levels: {\em observational},
{\em derivational},
and {\em validatory}.

   In the first, observational, case corpora - even processed
as simply as by concordance routines - can usefully display language
phenomena, both recording and drawing attention to them. This was one
of the earliest uses of IT for linguistic study, and remains important
though as corpora get larger it becomes harder to digest the concordance
information.

   Even at this level, however, there is the important issue of corpus
coverage versus representativeness. While one obvious use of corpora
is as a basis for grammars (Stubbs 1996),
they have become increasingly important for lexicographers
(see e.g. Thomas and Short 1996).
Here, while one function is to capture at least one
example of every configuration, word or word sense  (especially the latter),
another has been to display the relative frequency of lexical usage
(of value, for example, in building dictionaries for teaching). In both cases,
however, the issue of corpus representativeness arises
(Biber 1994, Summers 1996). What is
the corpus supposed to represent? And how do we know it is so
representative?

   There is a presumption, for some, that a large enough mass of
miscellaneous stuff taken from newspapers and so forth will be
representative of common, regular, or mainstream phenomena. However
it is more usual, as with the British National Corpus
(Burnard 1995), to develop
some set of selection criteria that draw on conventional or intuitively
acceptable notions of genre, and to gather samples of each. But this is a
far from scientific or rigorous basis for claims of proper status for
the resulting linguistic facts.

   At the same time, while even a simple concordance can be useful, IT
makes it possible to apply `low-level' linguistic processing of an
uncontroversial but helpful kind, for example lemmatization, tagging of
syntactic categories, labelling of local syntactic constituents (e.g. noun or
verb group) and even some marking of word senses (referring to some set of
dictionary senses).
Garside, Leech and Sampson (1987) illustrate both the possibilities
and the important contribution Lancaster has made here.
(It should be noted, however, that the opportunities
for analogous automatic processing of speech data, presuming the ability
to recognize and transcribe speech with reasonable accuracy, are currently
much more limited.)

   The second, derivative level of corpus use is potentially much more
interesting, but is also more challenging. It is foreshadowed by the
collection,
even at the first level, of simple frequency statistics, but is aimed at a
much more thorough analysis of data to derive patterns automatically:
lexical collocations, subcategorization behaviour, terminological
structure,
even grammar induction (Charniak 1993).
Such analysis presupposes first, some intuitive notion of the
name of the game as the basis for choosing both the primitive attributes
of the data
and the specification for the formal model of what is to be automatically
sought; and second, the actual algorithm for discovering model instances
in the data,
as indicated, for instance, by Gale, Church and Yarowsky (1994).
The problems here are challenging and are well illustrated by the attempt
to establish lexical fields objectively, by computation on data, rather
than by introspection supported by data inspection. Thus what features
of word behaviour in text are to be taken as the primitives for entity
description? What measures adopted to establish similarity of
behaviour both between a pair of words and, more importantly, over
a set of words to define a field, i.e. a semantic class? What operational
procedure will be applied to deliver and assess candidate classes? Cashing
in the notion of lexical field requires a whole formally and fully
defined discovery procedure, not to mention also some reasonable and
possibly automatic way of evaluating the definitions applied as
interpretations of the initial intuitive notion and, indeed, as
justifications for
the intuition itself.

   The potential value of IT for information extraction from large-scale
data bashing is obvious. But the difficulties involved, already indicated
for the determination
of lexical fields, are yet more evident in the idea
of deriving the genres,
even just for written discourse, of a language community
by operations on a (very) large neutral corpus, say the entire
annual intake of a major copyright repository. Genre is a function of many
language factors - lexical, syntactic, semantic, pragmatic
(communicative context and purpose) and, also, actual subject matter;
so both specifying and applying the primitive attributes
through which discourse sets will be differentiated, and hence genres
defined, is clearly no simple matter. The example however also
illustrates the range of useful outputs such a process can in principle
deliver the linguist: not merely indicative sets of actual discourses, but
higher-order genre definitions based on class membership (by analogy
with centroid vectors), as well as genre labelling for words in the lexicon.

   The third level, theory validation, is where the two areas of IT utility
for linguistics overlap. IT in principle offers great opportunities here,
through making it possible to evaluate a theory of some linguistic
phenomenon in a systematic, i.e. objective and comprehensive, way against
some natural corpus. But what does it mean to test a theory against a
corpus, informatively and unequivocally? If I have some theory of the
nature of syntactic or semantic representation,  I can check it for
propriety and coverage using a corpus, by seeing whether I can provide
representations for all the sentences
in the corpus. However such a test, as in
other cases, is only a negative one. If processing succeeds, it tells
me that my theory holds for this data, but not that it is the only possible
or best theory. The obvious problems for theory evaluation are thus on the
one hand the adequacy of
the corpus, and on the other the explanatory adequacy of the
theory. Taking these points further, natural corpora may be dilute, with
a low incidence of test instances (e.g. occurrences of rare word senses
for a model of sense selection); ambiguous, offering only very weak support
for
a theory
because there are many alternative accounts of some phenomenon (e.g.
sense selection either through lexical collocation or world knowledge), and
opaque, too rich to allow sufficiently discriminating testing on some
submodel through the interaction effects between phenomena (e.g. syntax
and lexicon).

   More importantly, using IT to validate a theory against a corpus
requires an {\em automatic} procedure for theory application,
the major issue for the model research
considered in the next section.

   The points just made have referred to the analysis of running text
data. But there is also one important, special kind of corpus to which
increasing attention is being paid, namely that
represented by a lexicon. A lexicon may
be viewed as providing second-order data about language use, rather than
the first-order data given by ordinary discourse. While the information
supplied by a dictionary has the disadvantage that it embodies the
lexicographer's biases, it has the advantage of providing highly
concentrated information, often in a relatively systematic way that reflects
the application of a special-purpose sublanguage. Exploiting this
information may involve hairy conversions from typesetting tapes, as
well as the further regularization required to develop a so-called
lexical database. But it is then in principle possible to derive a
higher-level classificatory structure over words from the bottom-level
entries. Early ideas here are illustrated by Sparck Jones (1964/1986),
more recent by Boguraev and Briscoe (1989). Of course corpus analysis for
text and lexicon
can be brought together, for example to select a domain
sub-lexicon, which may be linked with the syntactic and semantic
preferences of a domain grammar that is grounded in the text corpus.

\subsection*{Model research}

   The importance of IT for linguistic theory goes far beyond the
stimulus to model formation that browsing over volumes of data
may provide and even, though this is not to imply that such evaluation
is not of critical importance, beyond the testing of a theory against
a corpus. This is because, as mentioned earlier, computing offers not
only a natural context for the development and expression of {\em formal}
linguistic theories; it also places the most demanding, because 
of necessity principled,
requirements on theory, through theory application in systems
for implementing language-using tasks. This is not to imply that useful
systems cannot be built without theory, or at any rate without careful
and rigorous theory as opposed to some ad hoc application of some
plausible general idea. But the fact that NLP systems, for language
{\em interpretation} or {\em generation}
for some purpose, can be built is both
a challenge for, and a constraint on, those concerned with linguistic
theory.

   There are indeed several specific benefits for linguistics from
IT here.

   In relation to IT as a stimulus to formal model development, the
most extreme position is that the style of formal language theory that
computer science has also stimulated and enriched is the right kind
of apparatus for the formal characterization of natural languages
(see for example Gamut 1991). This
is a complicated matter because programming languages gain their special
power from eschewing the ambiguity that characterizes natural language.
However as computing systems have become more complex, computer science
theory has been obliged to seek a subtler and richer expressivity
(for example in capturing temporal phenomena), and thus might possibly
provide the means for characterising our language without damaging
over-simplification. The crux here is thus whether computer science
offers well-founded ways of cashing in the computational metaphor now
common, in both vulgar and philosophical parlance, for human activities
including the use of language.

   This still leaves open, however, both competence and performance-oriented
approaches. Thus taking language production as an example, we
can have both a formal, computational, {\em competence} theory
characterising
a syntactic model that would hopefully generate all and only the
syntactically legitimate sentences of a language. Or we can have
a formal, computational {\em performance} theory intended to model the way
humans actually go about producing syntactically acceptable strings.
Such a theory could indeed in principle encompass performance in the
behavioural limit by including e.g. mechanisms for restarting sentences
under certain production conditions. Thus because computation is essentially
about actually, as oppposed to possibly, doing things, it invites an
attack on flowing rather than frozen language.
Dowty, Karttunen and Zwicky (1985) and Sowa (1984) illustrate the wide
range of possibilities for such performance modelling.

   The business of processing naturally leads to the second level of
IT relevance for linguistics, that associated with building IT systems
for {\em tasks}. The point
here is that such systems are not just ones capable of exercising
language-using {\em functions}, like interpreting and answering a question,
responding to a command, endorsing a statement, i.e. systems with the
necessary bottom-level capabilities for language {\em use}. Even here
such systems have taken a critical step beyond the treatment of language
as a matter of words and sentences, and an ability to handle forms like
interrogatives or imperatives as defining sentence types.
The absolutely minimal level of functionality is represented by what
may be called `checking' responses, for example to some question
by noting that it
is a question asking whether X or not, or to a statement by offering
a paraphrase. It is possible to view such a form of model evaluation
as purely linguistic and without any real invocation of communicative
purpose or utterance context, but with the advantage that the model
evaluation involved does not depend on inspecting model-internal
representational structures (parse trees, logical forms etc) for
plausibility, a very dubious way of validating representations
of language form or
meaning.

   But since language is used for communication, IT would seem
to have a more substantive role in model testing
even at the level of individual
functions, e.g. by answering a question rather than by merely reformulating
it in some operation defined by purely linguistic relationships. Answering
a question appears to imply that a fully adequate interpretation of the
question has been attained. Thus we may imagine, for example, some
`database' of information to which questions may be applied. But such
strategies for model evaluation are of surprisingly limited value both
because of the constraints imposed by whatever the example data is, and
because of the essentially artificial restrictions imposed by the treatment
of sentence (utterance) function independent of larger communicative
purpose and context. Even the idea of answering questions implies relations
between different sentence functions, and models that attempt to account
for anaphor, for example, invoke above-sentence discourse.
This is evident in both Gazdar and Mellish (1989)'s and Allen (1995)'s
treatments of computational processing, for example.

   NLP systems are built for tasks like translation, inquiry, or
summarising that go beyond sentence function by requiring accounts of
communication and discourse (and therefore typically also have
not only to address
a range of sentence functions but also themselves subsume different
tasks). In general, properly done and not in such limited
application domains as to justify wholesale simplification, task systems
exercise the ability to determine meaning from text, or to deliver text
for meaning. They thus constitute the best form of evaluation
for linguistic models.
They can do this for the competence-oriented linguist if
required. But their real value is in performance modelling:
what are the {\em processes} of sentence and discourse interpretation or
generation? More specifically, if language has `components': morphology,
syntax, semantics, pragmatics, the lexicon (and these also above
the sentence, in discourse grammars) how do these {\em interact} in
processing,
i.e. what is the processor's architecture in terms of control flow?
How do components impose {\em constraints} on one another?
Winograd (1972) amd Moore (1986) equally show, in different situations
and applying rather different ideas, how significant the issue of
processor architecture is.

   It is possible to address process for single components, for example
in whether syntactic parsing is deterministic (Marcus 1980).
But if IT offers, in
principle, the `best' form of testing for language models because it avoids
the danger
of pretending that humans can assess objects that are really inaccessible,
namely `internal' meaning representations, this is also the
toughest form
of testing, for two reasons. First, how to evaluate task performance,
given this is the means of model assessment: for example, how to rate
a summarising system when in general there is no one correct summary of
a text? While linguistics makes use of judgement by informants, e.g. (and
notoriously) about grammaticality,
informant judgements about system performance for complex tasks are
much harder to make and much less reliable;
but in a disagreeable paradox, human participation with
the system in some task, for example in reading a summary in
order to determine whether to proceed to the full underlying text, is
either too informal at the individual
level or
too rough when based on many user decisions, to be an informative
method of model
evaluation. This exacerbates the problem, for task systems that are
inevitably multi-component ones that
depend both on individual language facts in
the lexicon and on general rules, of assessing the validity of
model detail. It should also be recognized that task systems
normally require knowledge of the non-linguistic world
to operate, so attributing
performance behaviour to the properties of the linguistic, as opposed
to non-linguistic, elements of the system as a whole can be hard.
These challenges and complexities of evaluation are further explored in
Sparck Jones and Galliers (1996).

   But it is further the case that while task systems can in principle offer
a base for
the evaluation of linguistic theory, in practice they may be of much
more limited value, for two reasons. One is the `sublanguage' problem,
where tasks are carried out in particular application domains: this
makes them suspect as vehicles for assessing the putatively general
models that linguists seek. The other is that practically useful systems,
e.g. for translation, can be primarily triumphs of hackery, with little
or only the most undemanding underpinning from models, which makes their
contribution to model evaluation suspect unless, as discussed further later,
this is taken as a comment on the whole business of language modelling.
Nevertheless, the key role that computation offers research on models is in
forcing enough {\em specificity} on a theory for it to be programmed and
operationally applied in autonomous action: humans can rely on hand waving,
but machines can't.

\section{IT actualities}

   Now, having rehearsed the potential utilities of IT generally (and hence
also of computer science) for linguistics, we can ask: How far has
IT actually had any impact
on linguistics? Further, has any impact been direct, through
computationally-derived data, or through model validation? Or has it been
indirect, through the recognition of computational paradigms? In relation
to data this influence would be most clearly shown
by a respect for statistics, and in allowing that
language-using behaviour may be influenced by frequency. This last may seem
an obvious property of language,
but acknowledging the computational paradigm brings it into
the open. At the theory level, the computational paradigm focusses not
so much on rules - a familiar linguistic desideratum - as on rule
application. Even when computational work adopts a declarative rather
than procedural approach, concern is always with what happens when
declarations are executed and so, for example, with compositional
effects in sentence interpretation.

   Overall, though this is an informal judgement (and also an amateur
one by a non-linguist), the impact of IT on linguistics
as a whole has been light,
and more peripheral than substantive.
\footnote{Certainly if the evidence of the linguistics shelves in a
major Cambridge bookstore is anything to go by.}
I shall attempt to summarize
the relevant work, and identify its salient features, and then seek reasons
for the lack of impact and interaction.

\subsection*{Data exploitation}

   It is clear that natural corpora can supply test data bearing on all
of morphology, syntax, semantics, pragmatics, and the lexicon, though the
processing for this (and indeed the
linguistic knowledge presupposed in this processing)
can vary. For example while it is relatively easy for English simply
to pick up all the word tokens (though of course also names, misspellings
etc etc) in a corpus, this may be rather less useful in e.g. German, where
freely formed compounds may have to be deconstructed. Similarly, for those
interested in syntax, offloading some of the data assembly work to the
machine depends on bootstrapping via a surface parser, say. But setting
this aside, what work under the heading computational data exploitation
has actually been done?

   Corpus use at the lowest, observational level appears to be spreading
- indeed has been referred to as one of the fastest growing areas
of linguistics (cf Stubbs 1996), even if it is not yet widespread:
it is illustrated, for example, by past uses of the Brown or
Lancaster/Oslo-Bergen
Corpora, and the use that is beginning, especially by lexicographers,
of the British National Corpus
(Burnard 1995). It is hard to measure in any precise
way how valuable such browsing and observational use of simple word
concordance and frequency data is, but the fact that serious publishers
are willing to put money behind corpus construction suggests that, at
least by such `applied' linguists, corpora are seen as useful, even
essential.
The range of possible corpus uses for descriptive purposes by linguists in
general rather than lexicographers is well-illustrated by e.g. such
ALLC-ACH '96
Conference titles as `Collocation and the rhetoric of scientific ideas:
corpus linguistics as a methodology of genre analysis'; and see also
Stubbs (1996), Thomas and Short (1996). Corpora of a
relatively considered, rather than casually assembled, kind are also
becoming increasingly common through the efforts of organizations
like the Linguistic Data Consortium and several European groups.
These descriptive uses of corpora have also been taken further, via the
application of taggers and parsers (e.g. Garside, Leech and Sampson
1987,
Brill 1995), to gather information about syntactic
constructions or about words that is dependent on syntactic contexts.

   Processing in this way leads naturally to the derivational use of
corpora. It may, for instance, be exploited to establish preferences
between parsing paths
for NLP. This is one example of the increasing interest in
exploiting corpora at the derivational level, which also includes
analysis for such purposes as establishing
selection criteria for word senses and identifying
synonym sets. The range, both of techniques and data
types, to explore is well shown in {\em Computational Linguistics}
(1993) and Boguraev and Pustejovsky (1996). Much of this work is restricted
to finding
pairwise associations between `objects' and has not progressed to
full-scale classification, but is already showing its value.
It should also be noted that at both this
level and the next, corpus analysis can provide useful information about the
lexical and structural properties of particular language worlds e.g.
of financial news stories. Equally, derivational work on lexical
as well as text corpora is in progress, including work on multilingual
databases (Copestake {\em et al.} 1995).

   Finally, corpora have not only been used observationally
or derivationally: they have to
some extent been used to validate theory. This is often indirect, in
the sense that e.g. corpora have been used to test syntax analysers where
evaluation is of the parser used, or of the
specific grammar, rather than of the grammar type; but it still
constitutes model evaluation. While simply running a parser with
grammar against a corpus can be very instructive, performance
evaluation may also be by comparing output with the reference analyses
of a `treebank' (Marcus, Santorini and Marcinkiewicz 1993),
Black {\em et al.} 1996).

\subsection*{Theory development}

   Turning now to to language modelling, and starting with IT as a stimulus
to the development of formal theories, there has been work
on all aspects of language.

   There have been accounts of morphology, as by Kaplan and Kay (1994), and
Koskenniemi (1983);
views of grammar, with Lexical Functional Grammar (Kaplan and Bresnan
1982),
Generalized Phrase Structure Grammar (Gazdar {\em et al.} 1985),
Head Phrase Structure Grammar
(HPSG) (Pollard and Sag 1994) and Tree Adjoining
Grammar (Joshi 1987),
where the influence of the computational paradigm is not only in
formal definition but in a concern with computability in a real and
not merely notional sense;
work on semantics covers both lexical aspects - see for example
Saint-Dizier and Viegas (1995), and the representation and derivation
of semantic structure, say compositionally (e.g. Alshawi 1992).
In pragmatics there has also been work on computational implicature
(see Cohen, Morgan and Pollack 1990),
and computational, because essentially algorithmic, accounts of
discourse phenomena like the use of anaphoric expressions
and focussing
(e.g. Grosz, Joshi and Weinstein 1995):
the emphasis on mechanisms underlying anaphors and focus,
for instance, has helped to throw light on their roles.
Computational modelling of the structure of the lexicon is an
area of growing interest, extending from the form of individual
entries to the (inference-supporting) organization of a lexicon
as a whole (Briscoe, Copestake and de Paiva 1993, Pustejovsky 1995).

   In general, IT's distinctive orientation towards process, discussed
earlier, has stimulated both process-oriented views of grammar,
as in Shieber (1987), and an enormous amount of work on generic
parsing technologies, which can be seen as abstract performance
modelling: cf, for example, Gazdar and Mellish (1989) and Allen (1995).
One feature of this research has been attempts to capture
bidirectionality as an operational rather than purely formal
property of grammars. Alongside all of this, and again motivated by
computational principles, there has also been effort on generic
formalisms for the specification of grammars, e.g. PATR
(Shieber 1992), and for the definition
of lexical information, as in DATR (Evans and Gazdar 1996).

   Some of this work has been done in what may be called an academic
spirit and informally evaluated, in the style of mainstream
linguistics, by examples. However the implementational philosophy of
IT in general has also stimulated an interest in descriptive
coverage for the relevant language component, for instance of syntax, and in
the objective assessment of an abstract language model by automatic
processing with some instantiation of the model, as in sentence parsing
with some particular grammar. In this context it is worth noting that
the notion of test data can be extended to cover not only natural
corpora but also so-called test suites, specifically-constructed data sets
designed to optimize on discriminative power and focus in testing.
At the same time, grammar construction and testing tools have been
developed, as by Carpenter and Penn (1993), and Kaplan and Maxwell (1993).

   However the really important point about this work
as a whole is that it has been
closely tied to work on building systems for NLP tasks, like translation
or data extraction from text, as illustrated in
Grosz, Sparck Jones and Webber (1986) and Pereira and Grosz (1993).
The stress to which
computational models have been subjected by being adopted for real
systems has been of benefit to theory development; and the business of
building task systems, especially for dealing with the interpretation
or generation of dialogue or
extended text, has led directly to
attempts to provide rigorous and detailed
accounts of language `objects' and language-using
operations in two important areas. These are
of dialogue and discourse structure, as in the computational
refinement and application of Rhetorical Structure Theory
(compare Mann and Thompson 1988 and Moore 1995), and
the organization of {\em world knowledge} for application in conjunction
with purely linguistic information.
Building and using discourse models, where text and world interact
and where discourse referents including events are characterized, has
stimulated significant, concrete work
in the computational community on the treatment of important
language constructs, namely anaphoric and temporal expressions.
The issues that arise here, of
how to represent and reason with world knowledge as required for
language interpretation or generation have, as Sowa (1984) or Allen
(1995) illustrate, to be tackled by any system builder, and thus pervade
the NLP literature. But while this area is particularly
important because it addresses
what is properly a determinant of adequate linguistic
theory, it is one in which linguists have generally not been explicitly or
specifically interested.

\subsection*{General observations}

   Now when we look carefully at all this actual
IT-related work, which is indubitably
respectable and informative in itself, there are two significant points
about it. The first is that at the intellectually challenging
derivational and model validation levels of work with data, and even more in
all the research aimed at theory development, this is primarily done
by those who label themselves at least as Computational Linguists, and
perhaps as Language Engineers: that is by those whose who are
either committed
even as descriptive or theoretical linguists to a computational
perspective or by those engaged in practical NLP.
Thus innovative statistical corpus, and lexicon, work has been stimulated
by operational needs, including those for sublanguage grammars for
particular applications. The second is that
this work appears to have had little impact on the linguistic
community at large, even with the computer only in the role of humble
handmaiden. Those giving realistic, or real, computation a role
in work on language appear to be a distinct community, neither
influencing not interacting with the larger world of linguistics.

   The separation is of course not absolute: thus
Cole, Green and Morgan (1995) shows there is some contact between
the two sides, some work on morphology, for instance,
draws on computational sources (cf e.g.
Fraser and Corbett 1995),
and HPSG is a fruitful area of mutual influence.
However in general, even where there
appears to be connectivity, this is either a consequence of the
ineluctabilities of language facts or, as in the area of formal semantics,
is less attributable to the influence of IT than to pressure from a common
higher cultural authority, namely logic. Thus even if Kamp's Discourse
Representation Theory (Kamp 1981, Kamp and Reyle 1989) is
of increasing interest to computational linguists and even to those
building NLP systems, insofar as linguists also engage with it this
because it is part of a tradition, exemplified in the work of
Montague and Partee (Partee, ter Meulen and Wall 1990), that has been
one common element of linguistics at large.
This is indirectly illustrated by the {\em Handbook of Contemporary
Semantic Theory} (Lappin 1996), which has one computational
chapter out of twenty two.   At the same time, computational
metaphors, like online processing in psycholinguistics, all too often lack
substance of the kind needed to write a parsing program or to define an
architecture so as to deliver phone-by-phone flow of control in speech
understanding.

   Why, therefore, since IT in principle offers linguistics so much,
has it in practice contributed so little?

\section{Analysis}

\subsection*{Linguistics' influence on IT}

   So far, I have focussed on the contribution by IT to linguistics.
As an additional input, in trying to explain why IT appears to have had
so little influence on linguistics in general, it is useful to ask whether
linguistics has figured with those engaged in
computational work in the language area, and in particular has affected
those building NLP task systems, for translation, database access and
so forth. Of course, anyone building operational systems is bound to
use language objects like actual grammars and dictionaries.
The question is rather whether practitioners respond to the general style
of linguistic research (whatever that is) or adopt specific linguistic
theories, for example Chomsky's, which have dominated linguistics for the
last decades. It might be expected that if linguistic research were to
play a significant part in NLP, this would help to promote feedback from
NLP into linguistics generally.

   On the whole, the influence of linguistics,
and especially linguistic theory, on NLP has been slight, other
than in the shared area of formal semantics and in some rather
particular and local respects where
individual pieces of work
have been exploited (for example discussions of prepositions).
Moreover even if it takes time for linguists' work to have an impact,
as is interestingly shown by the slow spread of long-standing 
contributions from the Prague School
(Hajicova, Skoumalova and Sgall 1995),
the way Halliday's Systemic Grammar (Kress 1976) has been applied in
sentence and text generation
is the exception rather than the rule; and
indeed it is arguable that philosophy, in the shape of Grice's maxims,
has had more influence on NLP than linguistics proper.
Some of this discontinuity is,
regrettably, attributable to ignorance and laziness on the part of
those who build systems, fuelled by an assumption that linguists do things
so differently there is no point in checking what they say. There is a
good deal of what may be labelled `wheel rediscovery' where, after
computational practitioners have become alerted to some topic e.g.
pragmatics and discourse, have worked on it for a while and had, maybe,
some ideas, they have found that the linguists have been there before them
and have already
made some descriptive or analytic progress which could with
advantage be exploited.

   However, there are also more respectable reasons why those who might be
interested in applying the ideas and findings of linguistics
have not done so. One is
that these are in fact inapplicable because of fundamental
differences of paradigm.
This is well illustrated by work done in the past on applying Chomsky's
theories, where attempts to build transformational parsers were misguided
and unsuccessful, even if Government and Binding has fared a little better
(Stabler 1992).
The second. which is more likely to affect system developers, is linguists'
`selectivity': it is perfectly legitimate for the linguist to concentrate
on some particular feature of language, e.g. tense and aspect, or
nominalization and compound nominals, and to offer a potentially valuable
account of it, in isolation. But the system builder {\em cannot} leave
things in such isolation: he must, for instance, treat the parsing of compounds
as only one aspect
of sentence processing. Yet he often finds that he cannot just
plug a linguist's account of a phenomenon,
like compound interpretation, into some slot in his system:
it rests either on incompatible presuppositions about the rest of
language, or is essentially lonely as a cloud. The third reason is
simply that the linguists' work is not
carried through to the level of specificity where it can be taken to provide
even the beginnings of a grounding for programs. 
This is illustrated by Kintsch and van Dijk's approach to discourse
representation
(e.g. in van Dijk and Kintsch 1983)
where, for all the attraction of their ideas on the
formation of `summarising' structures and interest of their experiments
with human subjects, there is a huge gaping hole for anyone seeking to
cash in, at the level of specificity required for programs, on the notions
of propositional inference involved. The same holds for any
attempt to exploit Sperber and Wilson (1986)'s Theory of Relevance:
compare Ballim, Wilks and Barnden (1991), for instance.
But these problems also arise in much less obviously challenging areas
like syntax and the lexicon. Even where an explicitly formal
viewpoint has been adopted, the outcome may be more descriptive than
analytic, or a mere sketch with voids of raw canvas.

   The pragmatics case that Sperber and Wilson represent
nevertheless also draws attention to the limits on the
{\em potential} benefits to be gained from linguistics. Though
definitions of the scope of linguistics vary, linguists generally agree
in eschewing the non-linguistic world. Now while practically useful
linguistic systems can be built with very little reference to any kind
of world, or {\em domain}, model of what's out there to be, or being, talked
about, many tasks, of which information inquiry is one, do require such
domain models; and providing these is often the hardest part of
building useful systems.
These domain models require all of: generalizations about
the kinds of things there are in this world, particularities about
individual entities, and inference capabilities subsuming both the
types of reasoning allowed and search procedures for executing these
on the knowledge base. The need to engage in reasoning on world
knowledge in order to support
the interpretation of input discourse, to carry out
some consequent task activity, and to provide for
the generation of output discourse
places particular emphasis on the properties required of
{\em meaning representations}
so that this {\em operational} interaction between language
and thought can be effected. Those linguists, generally cognitive linguists,
willing to push far into this area, and of whom Lakoff and Jackendoff may
be taken as instances
(e.g. in Jackendoff 1994), neverthless fail to tackle the issue in a manner,
and at a level of detail, appropriate for system builders (even ones
willing to undertake a lot of hard graft themselves).

   This point suggests some of the reasons why, in turn, IT has had so
little impact on linguistics.

\subsection*{IT's influence on linguistics}

   Some of these reasons are good, others are bad. They apply primarily
to the more important area of model formation and evaluation.

   There are good reasons to do with principle, and also ones to do with
practice.

   First, at the level of principle, there are genuine (even if ultimately
metaphysical) differences of view about the scope of theories about
language, and on a narrower view, of the scope of linguistics as oppposed
to, say, philosophy.
Thus some linguists may argue that IT's
concerns with the connection between language processing and reasoning in
NLP systems should have no bearing on linguistics, though this is not
a reason for rejecting computational linguistics altogether.
Second, there may be different views of what may be called the
style of linguistics, according to the relative emphasis placed on
formal theories of language or on descriptive coverage of languages, even
where everyone would like to think that their work has {\em some}
theoretical underpinning. Certainly there are fashions, with theoretical
linguistics currently so dominant that comprehensive description has
little status.
On this view, while it is the business
of theorists to account for linguistic phenomena, this can perfectly
well be done by means of critical cases and illustrative examples,
supported by sensitive sampling: it is no concern of linguistic
theorists to engage in comprehensive grammar or lexicon writing. Again,
this differentiates linguistics more from those working on NLP than
from all of computational linguistics.

   As practical reasons, it is first the case that much of what
computational linguistics, or NLP, has been able to do so far is
rather,
even very, crude in relation to the linguistic data. IT has not
been able to capture many of the phenomena, and refined distinctions
(lexical, syntactic or semantic), that descriptive linguists have
noted and for which theoretical linguists seek (though not necessarily
with complete success so far) to account, for example
the subtleties of adverbials, or of register.
It is also the case that IT has been primarily devoted to English,
and otherwise concerned only with major languages including the main
European ones, Russian, Japanese and Chinese and, increasingly, Arabic,
but has paid no attention to the many languages, ranging from
Djerbal to
Huichol, that figure in linguistics. Moreover, as at least some
NLP systems are rather more hackwork than could be wished, there are
cases where linguists genuinely have nothing to learn from the IT side.

   It has also to be recognized that the arrogance so characteristic
of those connected with IT - the self-defined rulers of the modern world -
is not merely irritating in itself, it is thoroughly offensive when
joined to ignorance not only of language, but of relevant linguists'
work.

   So while IT claims to offer linguistics an intellectual resource,
especially through its methodology, it does not appear to demonstrate its
value convincingly to the linguistics community.

   But the potentialities in IT for linguistics that I presented earlier,
and the actualities I have illustrated, are both important enough to
suggest that the main reasons for the lack of linguistics interest in
IT, and the lack of computational linguists' influence on their
mainstream colleagues, are due primarily to bad rather than good reasons.
Again, these are at both principled and practical levels.

   In my view, {\em coverage} (as for a complete
syntactic grammar), {\em interaction} (as between syntax and semantics),
{\em process} (as in identifying the sense of a word in a sentence),
and {\em integration} (as in combining morphological, syntactic
and lexical evidence in processing operations) are all
proper concerns for linguistics, and there is
no proper justification for neglecting or excluding them. Not
being willing to learn from IT's fundamental grounding in process,
in particular, is
placing crippling limits on the power and interest of linguistic theories.
Thus the crux in computational language processing is in dealing with
the {\em ambiguity} - lexical, structural and referential - that is
a fundamental feature of language: to interpret linguistic utterances
the system must resolve ambiguity, and to generate effective utterances
the system must minimise ambiguity. Doing this requires precisely that a system
has coverage, manages interaction, and executes process to combine
different sorts of information. It is not possible, as is so often the
case in the linguistics literature, to focus only on one aspect of
language and ignore the others, except as a temporary strategy; nor
is it legitimate, as is also so often the case in the linguistics
literature, to take for granted that understanding of a sentence or
text which it is the whole object of the enterprise to achieve and explain.
Again, not taking computational output, delivered by an independent
black box, as a superior way of testing a theory seems deliberately
unscientific: what better way of evaluating an account of the
distinctions between word senses
than to see what happens when a translation program uses it?
(Indeed, testing by system performance is exciting as well as principled.)
But even the specificity required for computation, in itself, is an object
lesson for formal theoretical linguists.
Those linguists who reject the lessons of computational linguistics and
NLP are
thus also mistakenly, or wrongly, subverting their own cause.

   On the practical side, it is impossible not to conclude that many
linguists are techno- and logico-phobes. 
It is true that understanding logic, formalisms, and computational
concepts, requires training to which some may be unwilling to dedicate
themselves. But, less defensibly,
the business of working
out, in the necessary grinding detail, what a program should or does do
is so exhausting that it is easier to say that doing it is irrelevant.

\subsection*{Conclusion}

   It may be that, though it is hard to discern significant IT impact
on linguistics outside the area labelled `computational linguistics',
IT is now so generally pervasive that it has begun to invade linguistic
thinking. But it is doubtful, for example, whether Chomsky's
minimalist programme
(Chomsky 1995), which appears to invoke some notions also
encountered in computational linguistics, in fact demonstrates there is
any material influence from IT.

   However, as NLP is forced, by tackling some tasks like interactive
inquiry, to address topics like dialogue structure, and automatic
speech and language processing continue to make progress, often with
surprising success by alien means, as in the use of Hidden Markov
Models for speech `recognition', there is much for linguistics to gain
from looking both at how computation does things and at what it finds.

   It is something of a caricature to see those engaged with
computation as crass technocrats for whom the expression
``non-computational theory'' is an oxymoron, and linguists as toffee-nosed
snobs unwilling to inspect the rude mechanicals' cranks and levers, and
a huge chasm between the two. But there is a gap that deserves to be
bridged because for linguists, and especially theorists other than those
whose metaphysic is resolutely anti-computational in any sense whatever,
there is everything to be learnt from appreciating the distinctions
between assumed, ideal, and real computation.

\vspace{10mm}

\subsection*{Note}

   I am most grateful to Gerald Gazdar for comments on my draft.

\vspace{10mm}

\section*{References}

Allen, J. (1995), {\em Natural Language Understanding}, Second Edition,
(Palo Alto, CA: Benjamin/Cummings).

Alshawi, H. (1992) (ed.), {\em The Core Language Engine},
(Cambridge MA: MIT Press).

Ballim, A., Wilks, Y. and Barnden, J. (1991), `Belief Ascription,
Metaphor, and Intensional Identification', {\em Cognitive Science},
15 (1), 133-171.

Biber, D. (1994), `Representativeness in Corpus Design', in
Zampolli, Calzolari and Palmer (1994).

Black, E. {\em et al.} (1996), `Beyond Skeleton Parsing: Producing a
Comprehensive Large-Scale General-English Treebank with Full
Grammatical Analysis', {\em COLING-96, Proceedings of the 16th
International Conference on Computational Linguistics}, Copenhagen,
107-112.

Boguraev, B. K. and Briscoe, E. J. (1989) (eds.), {\em Computational
Lexicography for Natural Language Processing}, (London: Longman).

Boguraev, B. and Pustejovsky, J. (1996) (eds.), {\em Corpus Processing for
Lexical Acquisition}, (Cambridge, MA: MIT Press).

Booij, G, and Marle, J. van (1995) (eds.) {\em Yearbook of Morphology
1994}, Dordrecht: Kluwer.

Bresnan, J. (1982) (ed.), {\em The Mental Representation of Grammatical
Relations}, (Cambridge, MA: MIT Press).

Brill, E. (1995),
Transformation-Based Error-Driven Learning and Natural Language
Processing: A Case Study in Part-of-Speech Tagging',
{\em Computational Linguistics}, 21 (4), 543-565.

Briscoe, E. J., Copestake, A. and de Paiva, V. (1993) (eds),
{\em Inheritance, Defaults and the Lexicon}, (Cambridge:
Cambridge University Press).

Burnard, L. (1995), {\em Users Reference Guide for the British
National Corpus}, Oxford University Computing Services, Oxford.

Carpenter, R. and Penn, G. (1993), {\em ALE: The Attribute Logic Engine,
Version 2.0 - User's Guide and Software}, Laboratory for Computational
Linguistics, Department of Philosophy, Carnegie-Mellon University.

Charniak, E. (1993), {\em Statistical Language Learning}, (Cambridge, MA:
MITP Press).

Chomsky, N. (1995) {\em The Minimalist Programme}, (Cambridge, MA: MIT
Press).

Cohen, P. R., Morgan, J. and Pollack, M. E. (1990) (eds.),
{\em Intentions in Communication}, (Cambridge, MA: MIT PRess).

Cole, J., Green, G. M. and Morgan, J. (1995) (eds.) {\em Linguistics and
Computation}, CSLI Lecture Notes 52, (Stanford: Centre for the Study of
Language and Information).

{\em Computational Linguistics} (1993), Special Issues on Using
Large Corpora, I and II, 19 (1) and 19 (2), 1-177 and 219-382.

Copestake, A. {\em et al.} (1995),
`Acquisition of lexical translation relations from MRDS',
Machine Translation, 9, (183-219).

Dijk, T. A. van and Kintsch, W. (1983), {\em Strategies of Discourse
Comprehension}, (New York: Academic Press).

Dowty, D. R., Karttunen, L. and Zwicky, A. M. (1985) (eds.),
{\em Natural language parsing}, (Cambridge: Cambridge University Press).

Evans, R. and Gazdar, G. (1996), `DATR: A Language for Lexical Knowledge
Representation', {\em Computational Linguistics}, 22 (2), 167-216,

Fraser, N. and Corbett, G. (1995), `Gender, Animacy, and Declensional Class
Assignment: A Unified Account for Russian' in Booij and van Marle (1995).

Gale, W. A., Church, K. W. and Yarowsky, D. (1994), `Discrimination
Decisions for 100,000-Dimensional Spaces' in Zampolli, Calzolari and
Palmer (1994).

Gamut, L. T. F. (1991) {\em Logic, Language and Meaning}, 2 vols.
(Chicago: University of Chicago).

Garside, R. Leech, G. and Sampson, G. (1987) (eds.), {\em The
Computational Analysis of English}, (London: Longman).

Gazdar, G., Klein, E., Pullum, G. K. and Sag, I. A. (1985),
{\em Generalized Phrase Structure Grammar}, (Oxford: Blackwell).

Gazdar, G. and Mellish, C. (1989), {\em Natural Language Processing
in Prolog}, (Reading, MA: Addison-Wesley).

Groenendijk, J., Janssen, T. and Stokhof, M. (1981) (eds.), {\em Formal
Methods in the Study of Language}, Mathematical Centre Amsterdam.

Grosz, B. J., Joshi, A. K. and Weinstein, S. (1995), `Centering:
a Framework for Modelling the Local Structure of Discourse',
{\em Computational Linguistics}, 21 (2), 203-225. 

Grosz, B. J., Sparck Jones, K. and Webber, B. L. (1986) (eds.),
{\em Readings in Natural Language Processing}, (Los Altos, CA:
Morgan Kaufmann).

Hajicova, E., Skoumalova, H. and Sgall, P. (1995),
`An Automatic Procedure for Topic-Focus Identification',
{\em Computational Linguistics}, 21 (1), 81-94.

Jackendoff, R. (1994), {\em Patterns in the Mind: Language and
Human Nature}, (New York: Harper Collins).

Joshi. A. K. (1987), `Introduction to Tree-Adjoining Grammars', in
Manaster-Ramer (1987), 87-114.

Kamp, H. (1981), `A Theory of Truth and Semantic Representation', in
Groenendijk, Janssen and Stokhof (1981), 277-322.

Kamp, H. and Reyle, U. (1989), {\em From Discourse to Logic}, 2 vols.
(Dordrecht: Kluwer).

Kaplan, R. M. and Bresnan, J. (1982), `Lexical-Functional Grammar', in
Bresnan (1982), 173-281.

Kaplan, R. M. and Kay, M. (1994), `Regular Models of Phonological
Rule Systems', with comments by M. Liberman and G. Ritchie,
{\em Computational Linguistics}, 20 (3). 331-380.

Kaplan, R. M. and Maxwell, J. (1993), `Grammar Writer's Workbench', 
ms, Xerox Palo Alto Research Centre, Palo Alto.

Koskenniemi, K. (1983), {\em Two-Level Morphology: A General
Computational Model for Word-Form Recognition and Production},
Publication 11, Department of General Linguistics, University
of Helsinki, Finland.

Kress, G. (1976), {\em Halliday: System and Function in Language},
(London: Oxford University Press).

Lappin, S. (1996) (ed.) {\em The Handbook of Contemporary Semantic
Theory}, (Oxford: Blackwell).

Manaster-Ramer, A. (1987) (ed.), {\em The Mathematics of Language},
(Amsterdam: J Benjamins).

Mann, W. C. and Thompson, S. A. (1988), `Rhetorical Structure Theory:
Towards a Functional Theory of Text Organization', {\em Text: an
Interdiciplinary Journal for the Study of Text}, 8 (2), 243-281.

Marcus, M. P. (1980), {\em A Theory of Syntactic Recognition for
Natural Language}, (Cambridge, MA: MITP Press).

Marcus, M. P., Santorini, B. and Marcinkiewicz, M. (1993), `Building a
Large Annotated Corpus of English: the Penn Treebank', {\em Computational
Linguistics}, 19 (2), 313-330.

Moore, J. D. (1995), {\em Participating in Explanatory Dialogues},
(Cambridge, MA: MIT Press).

Partee, B. H., ter Meulen, A. and Wall, R. E. (1990) {\em Mathematical Methods
in Linguistics}, (Dordrecht: Kluwer).

Pereira, F. C. N. and Grosz, B. J. (1993) (eds.), Special Volume:
Natural Language Processing, {\em Artificial Intelligence} 63 (1-2),
1-492.

Pollard, C. and Sag, I. A. (1994), {\em Head-Driven Phrase Structure Grammar},
(Chicago: University of Chicago Press).

Pustejovsky, J. (1995) {\em The Generative Lexicon}, (Cambridge, MA:
MIT Press).

Saint-Dizier, P. and Viegas, E. (1995) {\em Computational Lexical
Semantics}, (Cambridge: Cambridge University Press).

Shieber, S. M. (1987), {\em An Introduction to Unification-Based Approaches
to Grammar}, (Chicago: University of Chicago Press).

Shieber, S. M. (1992), {\em Constraint-Based Grammar Formalisms},
(Cambridge, MA: MIT PRess).

Sowa, J. F. (1984), {\em Conceptual Structures}, (Reading, MA:
Addison-Wesley).

Sparck Jones, K. (1986), {\em Synonymy and Semantic Classification},
Thesis, 1964 (Edinburgh: Edinburgh University Press).

Sparck Jones, K. and Galliers, J. R. (1996), {\em Evaluating
Natural Language Processing Systems}, Lecture Notes in Artificial
Intelligence 1083, (Berlin: Springer).

Sperber, D. and Wilson, D. (1986), {\em Relevance}, (Oxford: Blackwell).

Sperberg-McQueen, C. M. (1994), `The Text Encoding Initiative', in
Zampolli, Calzolari and Palmer (1994), 409-428.

Stabler, E. P. Jr (1992), {\em The Logical Approach to Syntax},
(Cambridge, MA: MIT Press).

Stubbs, M. (1996), {\em Text and Corpus Analysis}, (Oxford: Blackwell).

Summers, D. (1996), `Computational Lexicography:the Importance of
Representativeness in Relation to Frequency', in Thomas and Short (1996).

Thomas, J. and Short, M. (1996) (eds.), {\em Using Corpora for Language
Research}, (London: Longmans).

Winograd, T. (1972), {\em Understanding Natural Language}, (Edinburgh:
Edinburgh University Press).

Zampolli, A., Calzolari, N. and Palmer, M. (1994) (eds.),
{\em Current Issues in Computational Linguistics: In Honour of Don Walker},
(Dordrecht: Kluwer).

\end{document}